# GENETIC ALGORITHM FOR OPTIMAL DISTRIBUTION IN CITIES

Esteban Quintero
Universidad Eafit
Medellín, Colombia
equinterom@eafit.edu.co

Mateo Sánchez
Universidad Eafit
Medellín, Colombia
msanchezt@eafit.edu.co

Nicolas Roldan
Universidad Eafit
Medellín, Colombia
nroldanr@eafit.edu.co

## 1. ABSTRACT

The problem to deal with in this project is the problem of routing electric vehicles, which consists of finding the best routes for this type of vehicle, so that they reach their destination, without running out of power and optimizing to the maximum transportation costs. The importance of this problem is mainly in the sector of shipments in the recent future, when obsolete energy sources are replaced with renewable sources, where each vehicle contains a number of packages that must be delivered at specific points in the city , but, being electric, they do not have an optimal battery life, so having the ideal routes traced is a vital aspect for the proper functioning of these. Now days you can see applications of this problem in the cleaning sector, specifically with the trucks responsible for collecting garbage, which aims to travel the entire city in the most efficient way, without letting excessive garbage accumulate.

**CCS Concepts**

• **Information systems**➝Database management system engines • **Computing methodologies**➝Massively parallel and high-performance simulations.

**Keywords**

Algorithm; heuristics; problem of the traveling salesman; iterations.

## 2. INTRODUCTION

The accelerated development of new technologies has brought with it the problems involved in using them in the best way, without wasting their potential and taking into account all the limitations they bring. The situation with electric vehicles is not different, they have a very large field of action and bring a significant improvement, both to society and the environment, but these mean a shorter duration of each trip due to the limitations brought using electric batteries instead of fuels. This document will treat with different possible solutions that can reduce this problem as much as possible, through algorithms that analyze in different ways different data structures in which the information of each "map" will be stored.

**PAGE SIZE**

All material on each page should fit within a rectangle of 18 × 23.5 cm (7" × 9.25"), centered on the page, beginning 1.9 cm (0.75") from the top of the page and ending with 2.54 cm (1") from the bottom. The right and left margins should be 1.9 cm (.75").

The text should be in two 8.45 cm (3.33") columns with a .83 cm (.33") gutter.

## 3. SIMILAR PROBLEMS

### 3.1 Ant Colony Optimization (ACO)

The algorithm of the colony of ants is an algorithm that aims to mimic the behavior of these insects, which move from a beginning node to an end leaving a path, which over time is fading, to be followed by the other ants find him as the road disappears, the longer paths from the beginning node to the end will be forgotten over time, while for the short and efficient ones, the ants will reinforce it and continue using it.

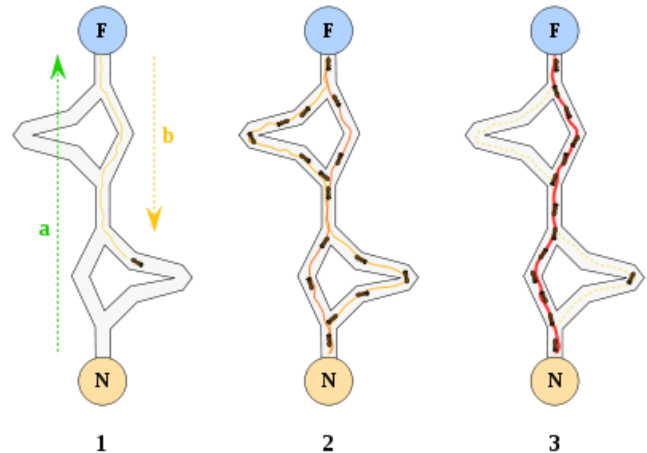

### 3.2 Genetic Algorithms

Genetic algorithms are a type of algorithms that, as the name says, "evolve" to find the best solution to a problem. These algorithms work by sending them a number of inputs, and, from them, generate a random output. After many outputs, the algorithm chooses the ones that have provided the best results, combining and altering them, in order to continue testing outputs and mixing solutions, until a sufficiently good result is reached that provides the optimal solution to the problem.

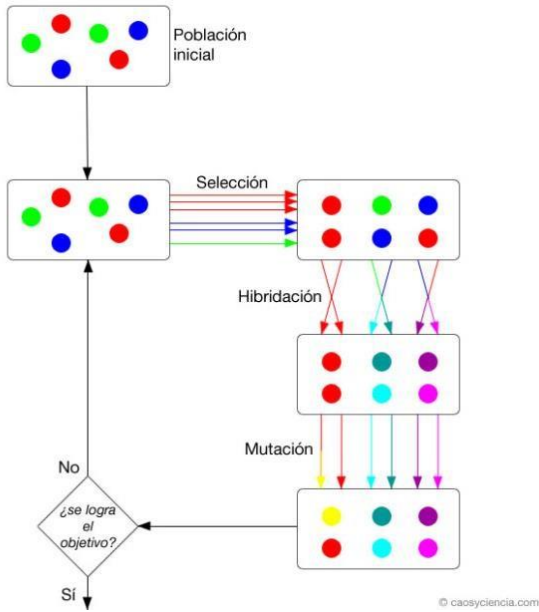

### 3.3 Constructive heuristics

A constructive heuristic is an algorithm that is building a complete solution from an empty solution adding to the latter, in each iteration, the best local choice of a set of possible choices. This method has been used to solve problems such as the problem of the traveling salesman, however, despite finding a complete solution, this is not the most effective. Next, an example of constructive heuristics is shown, always choosing the shortest arch to an unvisited node that leaves a node:

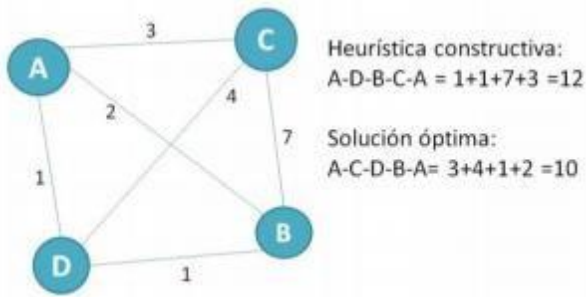

### 3.4 Taboo search

Created by Fred W. Glover, it is a mathematical optimization method that iteratively generates different solutions and stores them in a memory structure until a certain stop condition is met, and at the end, defines the final solution as the optimal one of the generated ones. For example, in the problem of the traveling salesman, a solution is generated from a constructive heuristic described above and from this, new solutions are generated by randomly exchanging the order in which cities are visited until they meet a certain number of iterations.

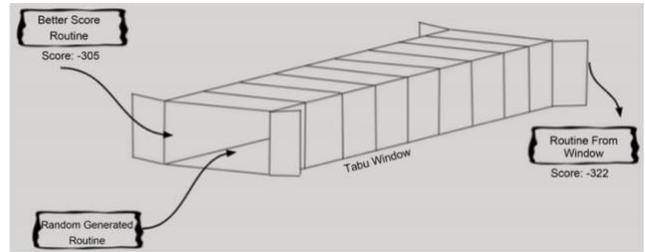

## 4. GENETIC ALGORITHM

These algorithms evolve a population of individuals subjecting it to random actions similar to those that act in biological evolution, as well as to a selection according to some criterion, depending on which it decides which are the most adapted individuals, that survive, and which are the least apt, which are discarded.

### 4.1 Data Structure

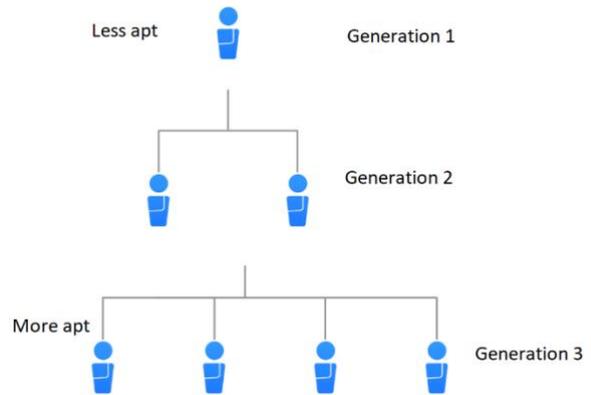

## 4.2 Operation of Data Structure

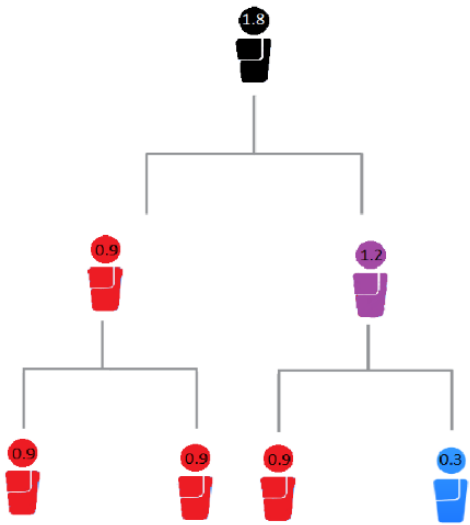

*The values represent the fitness of the individuals.

## 4.3 Design criteria of the data structure

Some of the criteria on which we rely is that this type of algorithms operate simultaneously with several solutions, instead of working sequentially as traditional techniques, it is also very easy to execute them in modern massively parallel architectures. In addition when they are used for optimization problems to maximize an objective function, they are less affected by local maxima (false solutions).

| Data | Time |
|---|---|
| 1.txt | 35.655 |
| 2.txt | 35.6 |
| 3.txt | 35.248 |
| 4.txt | 35.972 |
| 5.txt | 34.457 |
| 6.txt | 36.614 |
| 7.txt | 37.889 |
| 8.txt | 36.744 |
| 9.txt | 38.323 |
| 10.txt | 36.733 |
| 11.txt | 35.72 |

## 5.　CONCLUSIONS

In conclusion we see that is a problem that we will have to deal with sooner or later, and it may be earlier than imagined, that is why we must work on alternative solutions to deal with the problems that are to come.

The most important stuff that we learned from this solution is the same ones listed before in the criteria, some which sow that this is a very optimal solution.

## 5.1 Future work

In further works we would like to start working earlier so we would have time to develop other data structures to compare results.

## 5.2 Acknowledgements

We would like to thank our teachers, EAFIT University, and our classmates for helping us develop this project.